\pdfoutput=1

\documentclass[11pt]{article}

\usepackage{acl}

\usepackage{times}
\usepackage{latexsym}
\usepackage{booktabs}
\usepackage{graphicx}
\usepackage{amssymb}
\usepackage{xspace}
\usepackage{multirow}
\usepackage{amsmath}
\usepackage{bm}
\usepackage{listings}
\usepackage{bbm}
\usepackage{bbold}
\usepackage{algorithm2e}
\usepackage{csquotes}
\usepackage{enumitem}
\usepackage{pgfplots}
\usepackage{xurl}
\usepackage{titlesec}

\SetKwComment{Comment}{/* }{ */}
\usepackage[T1]{fontenc}

\usepackage[utf8]{inputenc}

\usepackage{microtype}

\newcommand{\event}[1]{\textit{#1}}

\renewcommand{\tt}[1]{\fontfamily{cmtt}\selectfont #1}

\newcommand{\eg}{\hbox{\emph{e.g.,}}\xspace}

\newcommand\vv{\ensuremath{\bm{v}}\xspace}

\newcommand\deberta{\textsc{DeBERTa}\xspace}
\newcommand\unlinkable{\textsc{DeBERTa-ul}\xspace}

\newcommand{\model}[1]{\textcolor{black}{\ensuremath{\mathcal{M}_{#1}}}}
\newcommand{\embed}[1]{\textcolor{black}{\ensuremath{\bm{e}_{#1}}}}

\newcommand{\newsz}[1]{\textcolor{black}{#1}}
\newcommand{\newhz}[1]{\textcolor{black}{#1}}

\newcommand*\samethanks[1][\value{footnote}]{\footnotemark[#1]}

\titlespacing{\paragraph}{%
  0pt}{
  0.0\baselineskip}{
  1em}
  
\title{Show Me More Details: \\Discovering Hierarchies of Procedures from Semi-structured Web Data}

\author{Shuyan Zhou$^\spadesuit$\thanks{~~Equal contribution.}, \quad
  Li Zhang$^\clubsuit$\samethanks, \quad
  Yue Yang$^\clubsuit$, \quad
  Qing Lyu$^\clubsuit$, \quad \\
  \textbf{Pengcheng Yin$^\spadesuit$,} \quad
  \textbf{Chris Callison-Burch$^\clubsuit$,} \quad
  \textbf{Graham Neubig$^\spadesuit$} \\
  $^\spadesuit$Carnegie Mellon University\quad\quad $^\clubsuit$University of Pennsylvania \\
  {\tt \{shuyanzh,pcyin,gneubig\}@cs.cmu.edu} \\ \tt{\{zharry,yueyang1,lyuqing,ccb\}@seas.upenn.edu}
}
  
\begin{document}
\maketitle
\begin{abstract}
Procedures are inherently hierarchical. To \event{make videos}, one may need to \event{purchase a camera}, which in turn may require one to \event{set a budget}. While such hierarchical knowledge is critical for reasoning about complex procedures, most existing work has treated procedures as shallow structures without modeling the parent-child relation.
In this work, we attempt to construct an open-domain hierarchical knowledge-base (KB) of procedures based on wikiHow, a website containing more than 110$k$ instructional articles, each documenting the steps to carry out a complex procedure.
To this end, we develop a simple and efficient method that links steps (\eg~\event{purchase a camera}) in an article to other articles with similar goals (\eg~\event{how to choose a camera}), recursively constructing the KB.
Our method significantly outperforms several strong baselines according to automatic evaluation, human judgment, and application to downstream tasks such as instructional video retrieval.\footnote{A demo with partial data can be found at \\ \url{https://wikihow-hierarchy.github.io/}. The code and the data are at \url{https://github.com/shuyanzhou/wikihow_hierarchy}.}

\end{abstract}

\begin{figure*}[t]
 \centering
 \includegraphics[width=1.0\textwidth]{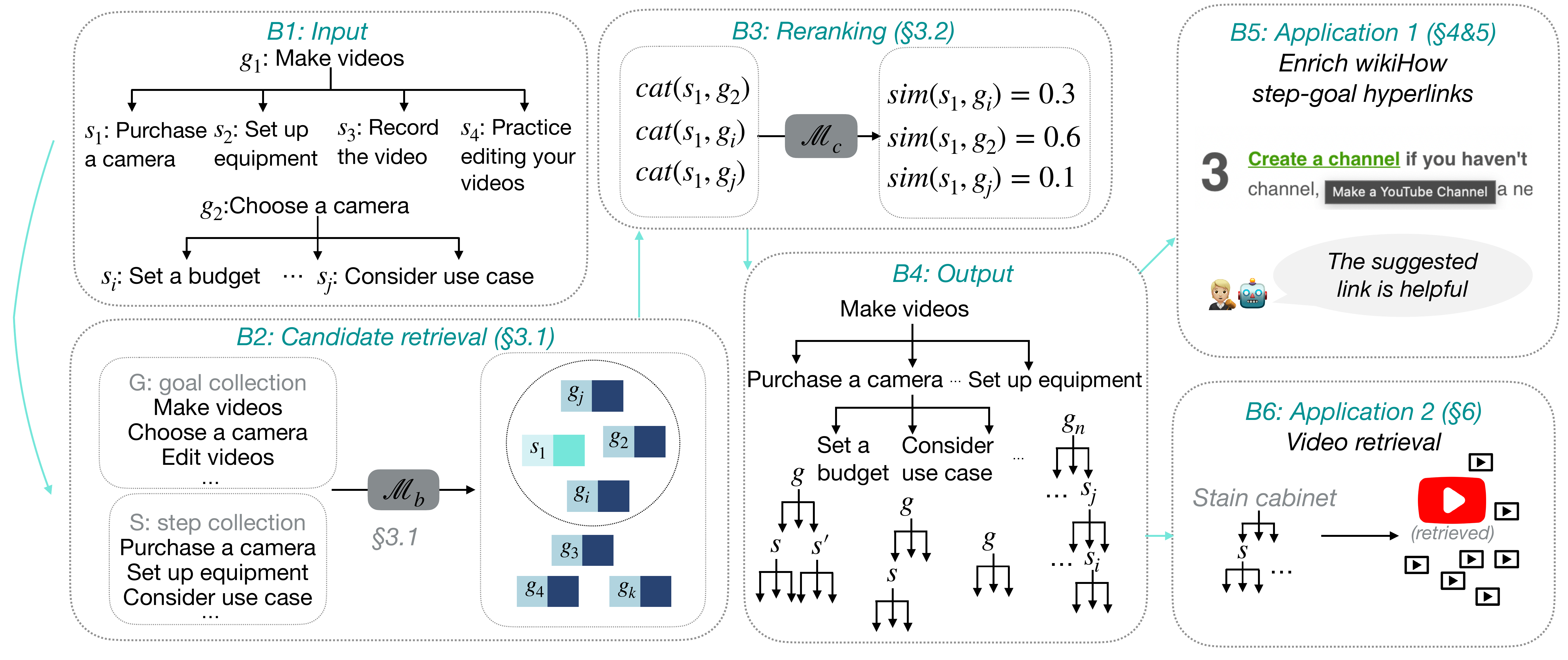}\\
 \caption{The overview of our proposed method. The input (\underline{B}lock1) and output (B4) of the hierarchy discovery model (B2, B3) and the applications (B5, B6) of the hierarchical knowledge base.
 }
 \label{fig:framework}
\end{figure*}
 
\section{Introduction}
A procedure includes some \textit{steps} needed to achieve a particular \textit{goal} \cite{momouchi-1980-control}. Procedures are inherently hierarchical: a high-level procedure is composed of many lower-level procedures. For example, a procedure with the goal \event{make videos} consists of steps like \event{purchase a camera}, \event{set up lighting}, \event{edit the video}, and so on, where each step itself is a procedure as well. Such hierarchical relations between procedures are recursive: the lower-level procedures can be further decomposed into even more fine-grained steps: one may need to \event{arrange the footage} in order to \event{edit the video}. 

Relatively little attention has been paid to hierarchical relations in complex procedures in the field of NLP.
Some work performed a shallow one-level decomposition and often required costly resources such as human expert task-specific annotation~\citep{chu2017distilling,zhang2020analogous, zhang2021learning}.
More attention has been paid in fields adjacent to NLP.  For example, \citet{lagos2017enriching} and \citet{pareti2014integrating} both create hierarchical structures in how-to documents by linking action phrases in one procedure to another procedure or by linking steps in how-to articles to resources like DBPedia~\cite{auer2007dbpedia}. This kind of linking is helpful for explaining complex steps to readers who do not have prior knowledge of the topic being explained.

In this paper, we revisit this important but understudied task to develop a simple and effective 
algorithm (\autoref{fig:framework}) to construct a hierarchical knowledge-base (KB) for over 110$k$ complex procedures spanning a wide range of topics from wikiHow, a large-scale how-to website that has recently become a widely-used resource in NLP \cite{zhou-etal-2019-learning-household,zellers-etal-2019-hellaswag,zhang-etal-2020-reasoning,zhang-etal-2020-intent}.\footnote{\url{www.wikihow.com}}
From each wikiHow article which represents a procedure, we follow \citet{zhang-etal-2020-reasoning} and extract the title as the goal~(\eg~$g_1$ in \autoref{fig:framework}), and the paragraph headlines as steps~(\eg~$s_1 \ldots s_n$).
Next, we decompose the steps by linking them to articles with the same or a similar goal~(\eg~$s_1$ to $g_2$). The steps of the linked article are treated as the finer-grained steps ($s_i$ to $s_j$) of the linked step~(s1). In this way, the procedural hierarchies go from shallow~(B1) to deep~(B4).  

To link steps and article goals, we employ a retrieve-then-rerank approach, a well-established paradigm in related tasks~\cite{wu2019zero,humeau2019poly}. Our hierarchy discovery model~(\S\ref{sec:model}) first \textit{independently} encodes each step and goal in wikiHow and searches the $k$ nearest goals of similar meaning for each step~(B2). 
Then, it applies a dedicated \textit{joint} encoder to calculate the similarity score between the step and each candidate goal, thus reranking the goals~(B3). 
This pipeline can efficiently search over a large candidate pool while accurately measuring the similarity between steps and goals. With each step linked to an article goal, a hierarchical KB of procedures is thus constructed.
 
We evaluate our KB both intrinsically and extrinsically. Intrinsically, the discovered links can be directly used to complete missing step-goal hyperlinks in wikiHow, which have been manually curated~(B5). Our proposed method outperforms strong baselines (\eg \citet{lagos2017enriching}) according to both automatic and human evaluation, in terms of recall and usefulness respectively~(\S\ref{sec:auto_eval}, \S\ref{sec:human_eval}).
Extrinsically, we consider the task of retrieving instructional videos given textual queries. We observe that queries that encode deeper hierarchies are better than those that do not (\S\ref{sec:ex_eval}). This provides evidence that our KB can bridge the high-level instructions and the low-level executions of procedures, which is important for applications such as robotic planning.

\section{Problem Formulation}\label{sec:problem_formulation}
We represent a procedure as a tree where the root node $n$ represents a goal and its children nodes $\texttt{Ch}(n)$ represent the steps of $n$. 
We formulate the hierarchy discovery task as identifying the steps among $\texttt{Ch}(n)$ that can themselves be a goal of some other finer-grained steps~(sub-steps), which are inserted into the tree. 

While this formulation could potentially be used on any large collection of procedures, we specifically focus on wikiHow. As shown in B1 of \autoref{fig:framework}, each article comprises a goal ($g$), and a series of steps ($\texttt{Ch}(g)$). Therefore, each article forms a procedure tree of depth one. 

We denote the collection of all goals and steps in wikiHow as $G$ and $S$ respectively. Our hierarchy discovery algorithm aims to link a step $s_i \in S$ to a goal $g \in G$ such that $g$ has the same meaning as $s_i$. It then treats $\texttt{Ch}(g)$ as $\texttt{Ch}(s_i)$. Given that $g$ and $s_i$ are both represented by textual descriptions, the discovery process can be framed as a \textit{paraphrase detection task}. This discovery process can be applied recursively on the leaf nodes until the resulting leaf nodes reach the desired granularity, effectively growing a hierarchical procedure tree~(B4 of \autoref{fig:framework}). 

\section{Hierarchy Discovery Model}\label{sec:model}
For each of the 1.5 million steps in the wikiHow corpus, we aim to select one goal that expresses the same procedure as the step from over 110$k$ goals. We propose a simple and efficient method to deal with such a large search space through a two-stage process. First, we perform \emph{retrieval}, encoding each step and goal \emph{separately} in an unsupervised fashion and select the $k$ most similar goals for each step $s$. This process is fast at the expense of accuracy. Second, we perform \emph{reranking}, \emph{jointly} encoding a step with each of its candidate goals in a supervised fashion to allow for more expressive contextualized embeddings. This process is more accurate at the expense of speed, since calculating each similarity score requires a forward pass in the neural network. The goal with the highest similarity score is selected and the step is expanded accordingly, as in B4 of \autoref{fig:framework}. 

\subsection{Retrieval}

In the first stage, we independently encode each step $s \in S$ and goal $g \in G$ with a model \model{b}, resulting in embeddings \embed{s_1}, \embed{s_2}, ..., \embed{s_n} and \embed{g_1}, \embed{g_2}, ..., \embed{g_m}. The similarity score between $s$ and $g$ is calculated as the cosine similarity between \embed{s} and \embed{g}. We denote this first-stage similarity score as $\textrm{sim}_1(s, g)$. Using this score, we can obtain the top-$k$ most similar candidate goals for each step $s$, and we denote this candidate goal list as $\texttt{C}(s)=[g_1, ..., g_k]$.
To perform this top-$k$ search, we use efficient similarity search libraries such as FAISS~\cite{JDH17}. 

We instantiate \model{b} with two learning-based paraphrase encoding models. The first is the \textsc{SP} model~\cite{wieting19simple, wieting2021paraphrastic}, which encodes a sentence as the average of the sub-word unit embeddings generated by SentencePiece~\cite{kudo-richardson-2018-sentencepiece}. The second is \textsc{SBert}~\cite{reimers-gurevych-2019-sentence}, which encodes a pair of sentences with a siamese BERT model that is finetuned on paraphrase corpus. 
For comparison, we additionally experiment with search engines as \model{b}, specifically Elasticsearch with the standard BM25 weighting metric~\cite{10.1561/1500000019}. We index each article with its title only or with its full article. We also experiment with Bing Search API where we limit the search to wikiHow website only\footnote{\url{www.bing.com}}. The BM25 with the former setting resembles the method proposed by~\citet{lagos2017enriching}.

\subsection{Reranking}

While efficient, encoding steps and goals independently is likely sub-optimal as information in the steps cannot be used to encode the goals and vice-versa.
Therefore, we concatenate a step with each of its top-$k$ candidate goals in $\texttt{C}(s)$ and feed them to a model \model{c} that jointly encodes each step-goal pair. Concretely, we follow the formulation of~\citet{wu2019zero} to construct the input of each step-goal pair as:
\begin{center}
\texttt{[CLS]} \textit{ctx} \texttt{[ST]}  \textit{step} \texttt{[ED]} \textit{goal} \texttt{[SEP]}     
\end{center}
where \texttt{[ST]} and \texttt{[ED]} are two reserved tokens in the vocabulary of a pretrained model, which mark the location of the step of interest. \textit{ctx} is the context for a step (\eg~its surrounding steps or its goal) that could provide additional information. The hidden state of the \texttt{[CLS]} token is taken as the final contextualized embedding. The second-stage similarity score is calculated as follows:
\begin{equation}
\small
\label{eqn:rerank}
    \textrm{sim}_2(s, g_i) = \textrm{proj}(\model{c}(s, g_i)) + \lambda \textrm{sim}_1(s, g_i)
\end{equation}
where $\textrm{proj}(\cdot)$ takes an $d$-dimension vector and turns it to a scalar with weight matrix $W \in \mathcal{R}^{d \times 1}$, and $\lambda$ is the weight for the first-stage similarity score. Both $W$ and $\lambda$ are optimized through backpropagation (see more about labeled data in~\S\ref{sec:auto_data}). 

With labeled data, we finetune \model{c} to minimize the negative log-likelihood of the correct goal among the top-$k$ candidate goal list, where the log-likelihood is calculated as:
\begin{align}
\small
\label{eqn:loss}
\begin{split}
    ll(s, g_i)=-\log\left(\textrm{softmax}\left(\frac{\textrm{sim}_2(s, g_i)}{\sum_{g_j \in \texttt{C}(s)}\textrm{sim}_2(s, g_j)}\right)\right)
\end{split}
\end{align}
Compared to the randomly sampled in-batch negative examples, the top-$k$ candidate goals are presumably harder negative examples~\cite{karpukhin2020dense} and thus the model must work harder to distinguish between them. We will explain the extraction of the labeled step-goal pairs used to train this model in~\S\ref{sec:auto_data}. 

Concretely, we experiment with two pretrained models as \model{c}, specifically BERT-base~\cite{devlin2018bert} and \deberta-large finetuned on the MNLI dataset~\cite{he2020deberta}. We pick them due to their high performance on various tasks~\cite{zhang2019bertscore}. \footnote{\label{bertscore}
\url{https://cutt.ly/oTx5gMM}.
BERTScore measures the semantic similarity between a pair of texts, similar to the objective of our reranking.}

In addition, we consider including different \textit{ctx} in the reranking input. For each step, we experiment with including no context, the goal of the step, and the surrounding steps of the step within a window-size $n$ ($n$=1).  
\subsection{Unlinkable Steps}
Some steps in wikiHow could not be matched with any goal. Such steps are \emph{unlinkable} because of several reasons. First, the step itself might be so fine-grained that further instructions are unnecessary (e.g. \event{Go to a store}). Second, although wikiHow spans a wide range of complex procedures, it is far from comprehensive. Some goals simply do not exist in wikiHow. 

Hence, we design a mechanism to predict whether a step is linkable or not explicitly. More specifically, we add a special token \texttt{unlinkable}, taken from the reserved vocabulary of a pretrained model, as a placeholder ``goal'' to the top-$k$ candidate goal list $\texttt{C}(s)$, and this placeholder is treated as the gold-standard answer if the step is determined to be unlinkable. The similarity score between a step and this placeholder goal follows \autoref{eqn:rerank} and $\textrm{sim}_1(s, \texttt{unlinkable})$ is set to the lowest first-stage similarity score among the candidate goals retrieved by the first-stage model.
Accurately labeling a step as \texttt{unlinkable} is non-trivial -- it requires examining whether the step can be linked to any goal in $G$. Instead, we train the model to perform this classification by assigning \texttt{unlinkable} to steps that have a ground-truth goal but this goal does not appear in the top-$k$ candidate goal list. The loss follows~\autoref{eqn:loss}.  

\section{Automatic Step Prediction Evaluation}\label{sec:auto_eval}
To train our models and evaluate how well our hierarchy discovery model can link steps to goals, we leverage existing annotated step-goal links. 

\subsection{Labeled Step-goal Construction}\label{sec:auto_data}
In wikiHow, there are around 21$k$ steps that already have a hyperlink redirecting it to another wikiHow article, populated by editors. We treat the title of the linked article as the ground-truth goal for the step. For example, as in B5 of \autoref{fig:framework}, the ground-truth goal of the step \event{Create a channel} is \event{Make a Youtube Channel}. We build the training, development and test set with a 7:2:1 ratio. 

\subsection{Results}\label{sec:auto_result}
\begin{table}[t]
  \centering\small
  \begin{tabular}{l|ccc}
  \toprule
  \textbf{Model} & \textbf{R@1} &  \textbf{R@10} & \textbf{R@30}\\
  \midrule
  \textsc{SP} & 35.8 & 64.4 & 72.5 \\
  \textsc{SBert} & 30.6 & 53.3 & 63.4 \\
  BM25 (goal only) & 30.5 & 51.6 & 61.1 \\
  BM25 (article) & 9.3 & 35.3 & 49.2  \\
  Bing Search & 28.0 & 47.9 & -  \\
  \midrule
  \midrule
  BERT  & 50.7 & 69.4  & -\\
  \deberta & \textbf{55.4} & \textbf{71.9} & - \\
   \hspace{3mm}$-$ surr & 54.3 & 71.6 & -\\
   \hspace{3mm}$-$ goal & 55.0 & 71.5 & - \\
   \hspace{3mm}$-$ both & 52.4 & 71.0 & -\\
   \hspace{3mm}$+$ unlinkable & 50.4 & 71.6 & -  \\
   \hspace{6mm}$+$ $\lambda=0$ & 51.9 & 71.4 & -\\
  \bottomrule
  
  \end{tabular}
  \caption{The recall@$n$ for different models on the test set. The top half are with paraphrase retrieval only and the bottom half are with taking the top-30 candidate goals generated by the best model (\textsc{SP}) and adding the reranking model. The best performance recall is \textbf{bold}. ``surr'' denotes the surrounding steps of the query step.\footnotemark}
  \label{tab:auto_result}
\end{table}
\begin{table*}[t]
    \centering\small
      \resizebox{2\columnwidth}{!}{%
    \begin{tabular}{c|llll}
    \toprule
    & \textbf{Step} & \textbf{Retrieval Prediction} & \textbf{Reranking Prediction}~(GT) & \textbf{Context}\\
    \midrule
    \multirow{3}{*}{\textbf{C1}} &  Learn to chop properly &  Learn Editing	& Chop Food Like a Pro & Use a Knife \\
    \cmidrule(r){2-5}
    &  Acquire a bike & Get on a Bike & Buy a Bicycle & Commute By Bicycle \\
    \cmidrule(r){2-5}
    & Get some vinyl records & Cut Vinyl Records & Buy Used LP Records & Buy a Turntable \\
    \midrule
    \multirow{5}{*}{\textbf{C2}} & \multirow{2}{*}{Open your coordinates} & Read UTM & Find Your Coordinates & Find the End Portal \\
    & & Coordinates & in Minecraft & in Minecraft \\
    \cmidrule(r){2-5}
    & \multirow{3}{*}{Fill in sparse spots} & \multirow{3}{*}{Remove Set in Stains} & \multirow{3}{*}{Fill in Eyebrows} & Shape Eyebrows (\textit{g}) \\
    & & & & Trim your brows (\textit{surr})\\
    & & & & Use a clear gel to set (\textit{surr})\\
    \bottomrule
    
    \end{tabular}%
    }
    \vspace{-2mm}
    \caption{The main failure modes of the candidate retrieval model (\textsc{SP}) that could be recovered by the reranking model. \textbf{Step}: the query step; \textbf{Retrieval Prediction}: the top-1 prediction of the best retrieval model \textsc{SP}; \textbf{Reranking Prediction}: the top-1 prediction of the best reranking model DeBERTa, it is also the ground-truth goal. By default, the \textbf{Context} refers to the goal of the query step. The last case lists both goal~(\textit{g}) and the surrounding steps~(\textit{surr}). }
    \vspace{-2mm}
    \label{tab:intrinsic_case}
\end{table*}
\autoref{tab:auto_result} lists the recall of different models without or with the reranking. \newhz{Precision is immaterial here since each step has \textit{only one} linked article.}

\paragraph{Candidate Retrieval} The \textsc{SP} model achieves the best recall of all models, outperforming \textsc{SBert} by a significant margin. Models based on search engines with various configurations, including the commercial Bing Search, are less effective. In addition, BM25~(goal only), which does not consider any article content, notably outperforms BM25~(article) and Bing Search\footnotetext{We are unable to get the top-30 results from Bing search because the web queries only return top-10 search results.}, implying that the full articles may contain undesirable noise that hurts the search performance. This interesting observation suggests that while commercial search engines are powerful, they may not be the best option for specific document retrieval tasks such as ours.

\paragraph{Reranking} We select the top-$30$ candidate goals predicted by the \textsc{SP} model as the input to the reranking stage. The recall@$30$ of the \textsc{SP} model is 72.5\%, which bounds the performance of any reranker.\footnote{We only experiment with \textsc{SP} because it is the best retrieval model, providing a larger improvement headroom.} As seen in the bottom half of \autoref{tab:auto_result}, reranking is highly effective, as the best configuration brings a 19.6\% improvement on recall@$1$, and the recall@$10$ almost reaches the upper bound of this stage. 
We find that under the same configuration, \deberta-large finetuned on MNLI~\cite{he2020deberta} outperforms BERT by 1.7\% on recall@$1$, matching the reported trends from BERTScore.\footnotemark[5] 

To qualitatively understand the benefit of the reranker, we further inspect randomly sampled predictions of \textsc{SP} and \deberta. We find that the reranker largely resolves \textit{partial matching} problems observed in \textsc{SP}. As shown in \textbf{C1} of \autoref{tab:intrinsic_case}, \textsc{SP} tends to only consider the action~(\eg~learn) or the object~(\eg~bike) and mistakenly rank those partially matched goals the highest. In contrast, the reranker makes fewer mistakes. In addition, we observed that the reranker performed better on rare words or expressions. For example, as shown in the last column of \textbf{C1}, the reranker predicts that ``vinyl records'' is closely related to ``LP records'' and outputs the correct goal while \textsc{SP} could not. 

Second, we observe that the surrounding context and the goal of the query step are helpful in general. Incorporating both contexts brings a 3\% improvement in recall@$1$. While steps are informative, they could be highly dependent on the contexts. For example, some steps are under-specified, using pronouns to refer to previously occurring contents or simply omitting them. The additional information introduced by the context helps resolve these uncertainties. In the first example of \textbf{C2}, the context ``minecraft'' is absent in the query step but present in the goal of that step. Similarly, in the second example, the context ``eyebrows'' is absent in the query step but present in both the goal and the surrounding steps. 

Finally, adding unlinkable prediction harms the recall@$1$ due to its over-prediction of \texttt{unlinkable} for steps whose ground-truth goal exists in the top-$k$ candidate list. We also experiment with setting a threshold tuned on the development set to decide which steps are unlinkable, in which case the recall@$1$ degrades from 55.4\% to 41.9\%. Therefore, this explicit learnable prediction yields more balance between the trade-offs.
In~\S\ref{sec:human_eval}, we will demonstrate that this explicit unlinkable prediction is overall informative to distinguish steps of the two types through crowdsourcing annotations. We empirically find that setting the weight of $\textrm{sim}_1(s,g)$~($\lambda$) to $0$ is beneficial in the unlinkable prediction setting. 

\section{Manual Step Prediction Evaluation}\label{sec:human_eval}
The automatic evaluation strongly indicates the effectiveness of our proposed hierarchy discovery model. However, it is not comprehensive because the annotated hyperlinks are not exhaustive.
We complement our evaluation with crowdsourced human judgments via Amazon Mechanical Turk~(MTurk).


Each example of annotating is a tuple of a step, its original goal from wikiHow, and the top-ranked goal predicted by one of our models. 
For each example, we ask three MTurk  workers to judge whether the steps in the article of the linked goal are exact, helpful, related, or unhelpful with regard to accomplishing the queried step. 
Details about the task design, task requirements, worker pay, example sampling, etc. are in~\ref{sec:cs_details}. 

We select SP, \deberta, and \deberta with unlinkable prediction and $\lambda=0$~(\unlinkable) for comparison.
We attempt to answer the following questions. First, does the performance trend shown in automatic evaluation hold in human evaluation? Second, can the unlinkable predictions help avoid providing users with misleading information \cite{rajpurkar-etal-2018-know}?

\begin{figure}
    \centering
    \includegraphics[width=0.5\textwidth]{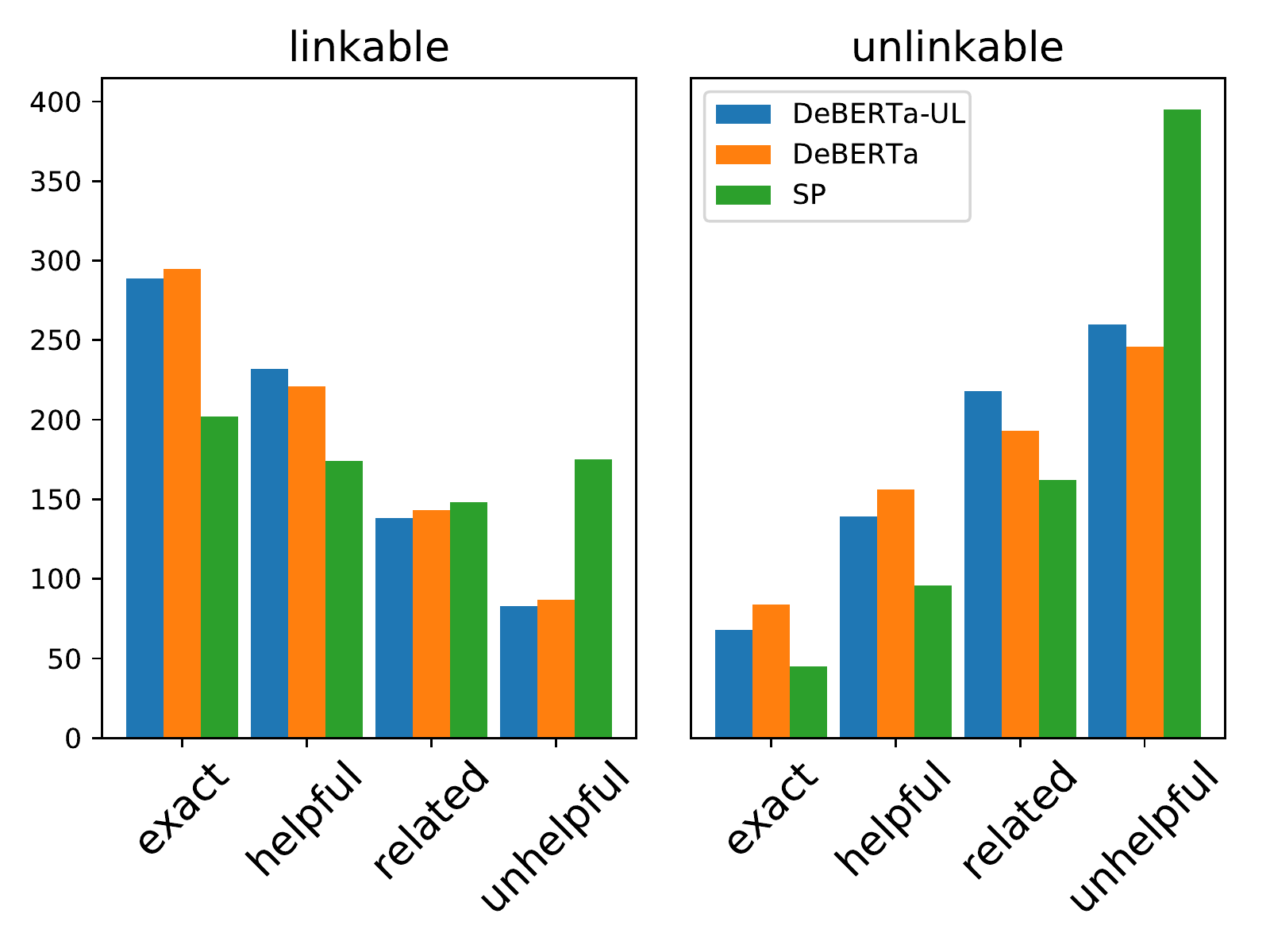}\\
    \vspace{-4mm}
    \caption{Crowd workers' ratings of step-goal links predicted by our models. The left graph shows steps linked to \textit{some} goals by the \unlinkable model, while the right shows steps those predicted as \texttt{unlinkable}.
    }
    \label{fig:crowdsourcing}
    \vspace{-4mm}
\end{figure}
For the purpose of the second question, 
we separate the examples into two groups. One contains linkable examples. Namely, those whose top-$1$ prediction is not predicted as \texttt{unlinkable} by the \unlinkable model. Ideally, the linked articles from these examples should be helpful. The other group contains unlinkable examples. For these, we evaluate the second-highest ranked prediction of the \unlinkable model. Ideally, the linked articles from these examples should be unhelpful.

The corresponding crowd judgment is shown in Figure~\ref{fig:crowdsourcing}. Comparing the models, the \deberta model and the \unlinkable model have similar performance, while greatly outperforming the SP model. This shows that our proposed model decomposes much more helpful finer-grained steps to assist users with tasks, similar to the trend observed in our automatic evaluation. Comparing the two graphs, it is apparent that when the \unlinkable model predicts \texttt{unlinkable} for a step, 
the suggested decompositions of all models are more likely to be unhelpful. This implies the high precision of the \texttt{unlinkable} prediction, effectively avoiding misleading predictions. Note that our study does not explicitly require subjects to carry out the task, but only annotates whether they find the instructions helpful.

\section{Application to Video Retrieval}
\label{sec:ex_eval}
In addition to intrinsic evaluation, 
we take a further step to study the usefulness of our open-domain hierarchical KB to downstream tasks. 
We select video retrieval as the extrinsic evaluation task, which aims at retrieving relevant how-to videos for a textual goal to visually aid users. More formally, given a textual goal $g$, the task is to retrieve its relevant videos $v_g$ from the set of all videos, with a textual query $q$. 
Intuitively, our KB can be useful because videos usually contain finer-grained steps and verbal descriptions to accomplish a task. Therefore, the extra information presented in decomposed steps could benefit retrieving relevant videos.

\subsection{Dataset Construction} \label{sec:dataset_construction}
We use Howto100M~\cite{miech2019howto100m} for evaluation. It is a dataset of millions of instructional videos corresponding to over 23$k$ goals.
We construct our video retrieval corpus by randomly sampling $1,000$ goals~(\eg~\event{record a video}) with their relevant videos. 
The relevant videos $\vv_{g}=\{v_1, v_2, ..., v_n\}$ of each goal $g$ in the dataset are obtained by selecting the top 150 videos among the search results of the goal on YouTube.\footnote{Although the relevance between a goal and a video is not explicitly annotated in the Howto100M dataset, we argue that with the sophisticated engineering of the YouTube video search API and hundreds of thousands user clicks, the highly ranked videos likely demonstrate the queried goal.}
For each goal $g$, we randomly split its relevant videos $\vv_{g}$ into three sub-sets $\vv_{g}^\textrm{tr}$, $\vv_{g}^\textrm{dev}$ and $\vv_{g}^\textrm{test}$ with a ratio of 7.5:1.25:1.25, as the training, development, and testing sets.\footnote{We explain more about the appropriateness of the downstream video retrieval task setup in \ref{sec:vr_split}.}

\begin{table}[t]\small
  \centering
  \resizebox{\columnwidth}{!}{%
  \begin{tabular}{l|ccccc}
  \toprule
  \textbf{Query} & \textbf{R/P@1} &  \textbf{R/P@10} & \textbf{R/P@25} & \textbf{R/P@50} & \textbf{MR}\\
  \midrule
  $\textsc{l}_{0}$ & 2.2/89.2 & 19.2/78.1 & 39.9/66.0 & 56.6/48.2  & 79.49 \\
  $\textsc{l}_{1}$ & 2.2/88.0 & 19.2/78.0 & 40.1/66.4 & 58.1/49.6 & 75.79 \\
  $\textsc{Fil-l}_{1}$ & 2.2/\textbf{89.9} & 20.2/81.7 & 43.1/71.2 & 63.2/53.8 & 66.32 \\
  $\textsc{Fil-l}_{2}$ & 2.2/89.4 & \textbf{20.3/82.7} & \textbf{43.9/72.3} & \textbf{65.0/55.2} & \textbf{63.38} \\
  \midrule
  \midrule
  $\textsc{l}_{0}$ & 12.1/81.7 & 59.8/42.8 & 71.9/20.8 & 77.9/11.3  & 41.60\\
  $\textsc{l}_{1}$ & 11.8/79.7 & 61.2/43.9 & 74.1/21.4 & 80.5/11.6  & 36.70 \\
  $\textsc{Fil-l}_{1}$ & 12.4/83.7 & 66.0/47.3 & 77.4/22.4 & 82.9/\textbf{12.0} & 33.35 \\
  $\textsc{Fil-l}_{2}$ & \textbf{12.5/84.4} & \textbf{66.1/47.7} & \textbf{78.0/22.5} & \textbf{83.3/12.0} & \textbf{32.30} \\
  \midrule
  \midrule
  $\textsc{l}_{0}$ & 11.4/82.6 & 59.2/45.2 & 71.8/22.1 & 77.8/12.0 & 43.11 \\
  $\textsc{l}_{1}$ & 11.2/81.3 & 60.4/46.2 & 73.8/22.7 & 79.9/12.3 & 38.19 \\
  $\textsc{Fil-l}_{1}$ & \textbf{11.7/85.1} & 64.8/49.5 & 77.2/23.8 & 82.2/\textbf{12.7} & 34.76\\
  $\textsc{Fil-l}_{2}$ & 11.6/84.5 & \textbf{65.5/50.0} & \textbf{77.9/24.0} & \textbf{82.7/12.7} & \textbf{34.13} \\
  \bottomrule
  
  \end{tabular}%
  }
  \caption{The Recall/Precision@$N$ (\%, $\uparrow$) and mean rank (MR, $\downarrow$) with different queries on the relevant video retrieval task on the training (top), development (middle) and the test set (bottom). The best performance on each set is \textbf{bold}.}
  \label{tab:vir_result}
  \vspace{-4mm}
\end{table}

\subsection{Setup}
Since our KB is fully textual, we also represent each video textually with its automatically generated captions.
For the search engine, we use Elasticsearch with the standard BM25 metric~\cite{10.1561/1500000019}.\footnote{We find the performance of a neural model (BERT finetuned on query/video caption pairs) significantly lower than BM25 and therefore, we only report the results with BM25.} We denote the relevance score calculated by BM25 between the query $q$ and a textually represented video $v$ as $\text{Rel}(q,v)$.

We experiment with four different methods, which differ in how they construct the query $q$:

\paragraph{$\textsc{l}_{0}$: Goal only.} The query is the goal $g$ itself. This is the minimal query without any additional hierarchical information. The relevance score is simply $\text{Rel}(q,v) = \text{Rel}(g,v)$.

\paragraph{$\textsc{l}_{1}$:~Goal~+~Children.} The query is a concatenation of the goal $g$ and its immediate children steps $\texttt{Ch}(g)$. 
This query encodes hierarchical knowledge that already exists in wikiHow. The relevance score is then defined as a weighted sum, $\text{Rel}(q,v) = w_g \text{Rel}(g,v) + w_{s} \sum_{s \in \texttt{Ch}(g)}  \text{Rel}(s,v)$. The weights $w_g$ and $w_{s}$ are tuned on a development set and set to 1.0 and 0.1 respectively.

\paragraph{$\textsc{Fil-l}_{1}$: Goal + Filtered children.} The query is a concatenation of the goal $g$ and a filtered sequence of its children $\texttt{Ch}(g)$. Intuitively, decomposing a goal introduces richer information but also introduces noise, since certain steps may not visually appear at all~(\eg~\event{enjoy yourself}). 
Therefore, we perform filtering and only retain the most informative steps, denoted by $\texttt{Ch}'(g)$.
Specifically, to construct $\texttt{Ch}'(g)$ for a goal $g$, we use a hill-climbing algorithm to check each step $s$ from $\texttt{Ch}(g)$, and include $s$ into the query only if it yields better ranking results for the ground-truth videos in the training set $\vv_g^\textrm{train}$.\footnote{See Algorithm \ref{alg:search} in Appendix for more details.} 
The relevance score is defined as $\text{Rel}(q,v) = w_g \text{Rel}(g,v) + w_{s} \sum_{s \in \texttt{Ch}'(g)}  \text{Rel}(s,v)$, where $w_g$ is set to 1.0 and $w_s$ is set to 0.5 after similar tuning. 

\begin{table}[t]
    \centering\small
    \resizebox{0.9\columnwidth}{!}{%
    \begin{tabular}{l|l}
    \toprule
    Goal & Stain Cabinet \\
    \midrule
    $\textsc{Fil-l}_{1}$ & Purchase some stain colors to test \\
    \midrule
    \multirow{4}{*}{$\textsc{Fil-l}_{2}$} & $\textsc{Fil-l}_{1}$ + \\
    & Buy cloth with which to apply the stain \\
    & Unscrew the cabinet from the wall \\
    & Clean your workspace \\
    \midrule
    \multirow{4}{*}{KM}& Remove the doors \\
    & Sanding the front \\
    & Top coat \\
    & Finished look \\
    \midrule\midrule
    Goal &  Make Avocado Fries \\
    \midrule
    \multirow{3}{*}{$\textsc{Fil-l}_{1}$} & Bake the avocado fries until they are golden \\
    & Dip the avocado wedges into the egg \\
    & and then the breadcrumbs \\
    \midrule
     \multirow{6}{*}{$\textsc{Fil-l}_{2}$} & $\textsc{Fil-l}_{1}$ +  \\
    & Preheat the oven \\ 
    & Peel and pit the avocados \\
    & Cut your avocado in half and remove the stone \\
    & Let rise \\
    & Finished, cool and enjoy \\
    \midrule
    \multirow{4}{*}{KM}& 2 large avocados ... \\
    & pinch of salt, pinch of pepper \\
    & two eggs, beaten ... \\
    & bake at 425F 20 min until golden bros ... \\
    \bottomrule
    \end{tabular}%
    }
    \vspace{-2mm}
    \caption{The queries and the key moments (KM) for two goals. ``...'' represents the omission of steps that describe the ingredients to save space. The first selected video is \href{https://www.youtube.com/watch?v=h9k0T25_NxA}{\texttt{h9k0T25\_NxA}} and the second is \href{https://www.youtube.com/watch?v=o7uVUmPph6I}{\texttt{o7uVUmPph6I}}.}
    \label{tab:vir_case}
    \vspace{-5mm}
\end{table}

\paragraph{$\textsc{Fil-l}_{2}$: Goal + Filtered children + Filtered grand-children.} The query is the concatenation of the goal $g$ and a filtered sequence of its immediate children $\texttt{Ch}(g)$ and grandchildren $\texttt{Ch}(s)$ ($s \in \texttt{Ch}(g)$). These filtered steps are denoted by $\texttt{Ch}'(g+\texttt{Ch}(g))$. This two-level decomposition uses the knowledge from our KB, therefore including lower-level information about the execution of the goal. We perform the same filtering algorithm as in $\textsc{Fil-l}_{1}$, and we define $\text{Rel}(q,v) = w_g \text{Rel}(g,v) + w_{s}\sum_{s \in \texttt{Ch}'(g+\texttt{Ch}(g))}  \text{Rel}(s,v)$. $w_g$ is set to 1.0 and $w_s$ is set to 0.5. 

\subsection{Results}

We report the precision@$N$, recall@$N$ and mean rank~(MR) following existing work on video retrieval~\cite{Luo2021CLIP4Clip} (see \S\ref{sec:eval_metrics} for metric definitions). \autoref{tab:vir_result} lists the results. 
First, queries that encode hierarchies of goals ($\textsc{l}_{1}$, $\textsc{Fil-l}_{1}$ and $\textsc{Fil-l}_{2}$) are generally more beneficial than queries that do not ($\textsc{l}_{0}$). The steps of goals enrich a query and assist the retrieval. Second, video-oriented filtering yields significant improvement over the un-filtered $\textsc{l}_{1}$ queries since it produces a set of more generalizable steps that are shared among multiple videos. Although steps in wikiHow articles are human-written, they are not grounded to real-world executions of that goal. 
Many steps do not have corresponding executions in the videos and become noisy steps in the $\textsc{l}_{1}$ queries. 
More interestingly, we observe that queries using deeper hierarchies ($\textsc{Fil-l}_{2}$) outperform the shallower ones~($\textsc{Fil-l}_{1}$) in most cases.
This is probably due to the fact that how-to videos usually contain detailed (verbal) instructions of a procedure, which are better aligned with more fine-grained steps found in $\textsc{Fil-l}_{2}$.

In our qualitative study, we investigate how $\textsc{Fil-l}_{2}$ queries with deeper hierarchies help retrieval. \autoref{tab:vir_case} list $\textsc{Fil-l}_{1}$ and $\textsc{Fil-l}_{2}$ queries for two goals. 
We find that the $\textsc{Fil-l}_{2}$ queries are more informative and cover more aspects. For example, the $\textsc{Fil-l}_{2}$ queries for \event{stain cabinet} and \event{make avocado fries} consist of the preparation, actual operations, and the post-processing steps, while the $\textsc{Fil-l}_{1}$ query only contains the first one. 
In addition, we search the goals on Google and list the key moments of some randomly sampled videos.\footnote{Key moments are either identified manually or are extracted automatically by YouTube. \url{https://cutt.ly/qTcxSi6}
} These key moments textually describe the important clips of the videos, and therefore they presumably also serve as the query for the goal. We find that the $\textsc{Fil-l}_{2}$ query of \event{make avocado fries} explains a few necessary steps to accomplish this goal, while the key moment is mostly composed of the ingredients of this dish. This comparison suggests the potential integration of our induced hierarchical knowledge to identify key moments in videos in the future.

\section{Decomposition Analysis}\label{sec:hierarchy_analysis}
\begin{figure}[t]
    \centering
    {
    \includegraphics[width=0.5\textwidth]{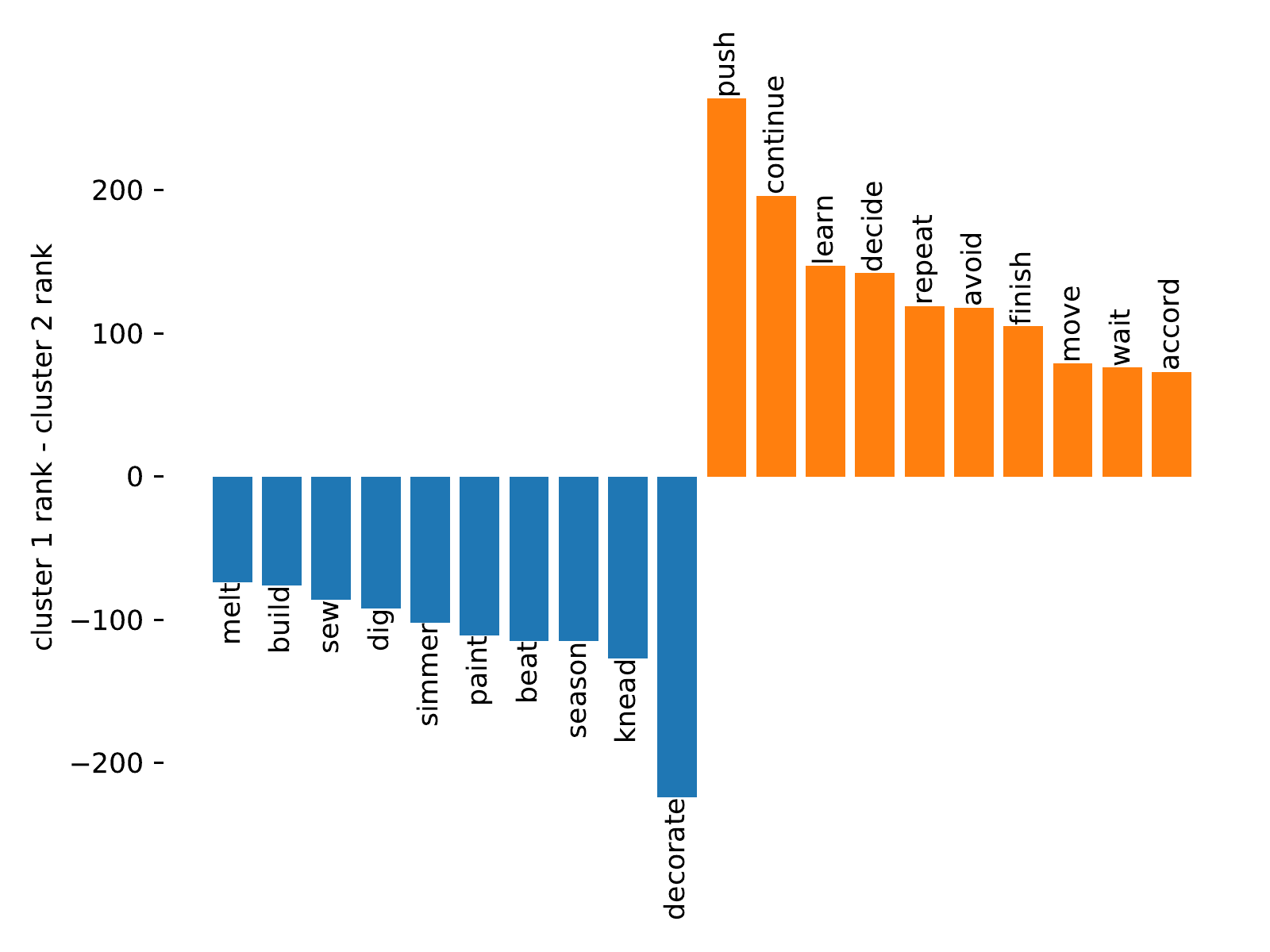}\\
    \vspace{-4mm}
    }
    \caption{The verbs with largest rank difference in two clusters. The blue bars are words becoming less frequent in cluster 2 (decomposed steps) and the orange bars are words becoming more frequent.}
    \label{fig:part_verb_distribution}
\end{figure}
In this section, we study the properties of the hierarchies. First, what kind of steps are likely to be linked to another goal and are thus decomposed? Second, what do the decomposed steps look like?

We group steps into two clusters. The first contains the immediate steps of a goal ($s \in \texttt{Ch}(g)$) whose prediction is not \texttt{unlinkable}. The second contains the decomposed steps of the steps in the first cluster ($s' \in \texttt{Ch}(s)$). We use spaCy~\cite{spacy} to extract and lemmatize the verb in each step and rank the verbs by their frequency in each cluster. 
Next, the top-$100$ most frequent verbs in each cluster are selected and we measure the rank difference of these verbs in the two clusters.
\autoref{fig:part_verb_distribution} plots the verbs with largest rank difference and the full figure is in \autoref{fig:full_verb_distribution}. 
We observe that verbs that convey complex actions and intuitively consist of many other actions become less frequent after the decomposition~(\eg decorate). 
On the other hand, verbs that describe the action itself gain in frequency after the decomposition~(\eg push, hold, press). 
This observation follows our assumption that the decomposition would lead to more fine-grained realizations of a complex procedure. 
Some other more abstract actions such as ``learn'' and ``decide'' also increase in frequency, as some low-level goals are explained with more complex steps.

\section{Related Work}
\paragraph{Linking Procedural Events} To the best of our knowledge, two other pieces of work \citet{pareti2014integrating,lagos2017enriching} tackled the task of linking steps in procedures to other procedures. Both of them also drew the procedures from wikiHow. While we share the same task formulation, our work makes several additional contributions: (1) a retrieval-then-rerank method significantly increases linking recall; (2) more comprehensive experiments with the manual and the downstream evaluation that showcases the quality and usefulness of the linked data and (3) experiments and data with broader coverage over all of WikiHow, not just the Computer domain.
\paragraph{Procedural Knowledge}
\newhz{Procedural knowledge can be seen as a subset of knowledge pertaining to \textit{scripts} \cite{abelson1977scripts, rudinger2015learning}, \textit{schemata} \cite{rumelhart1975notes} or events. A small body of previous work \cite{9070972} on procedural events includes extracting them from instructional texts \cite{10.1145/584955.584977,delpech-saint-dizier-2008-investigating, zhang-etal-2012-automatically} and videos \cite{alayrac2016unsupervised, yang2021induce}, reasoning about them \cite{takechi-etal-2003-feature,tandon-etal-2019-wiqa,rajagopal-etal-2020-ask}, or showing their downstream applications \cite{phdthesis,zhang-etal-2020-reasoning,yang-etal-2021-visual,zhang-etal-2020-analogous,lyu-etal-2021-goal}, specifically on intent reasoning \cite{DBLP:conf/aaai/SapBABLRRSC19,dalvi-etal-2019-everything,zhang-etal-2020-intent}.
Most procedural datasets are collected by crowdsourcing then manually cleaned \citep{singh2002open, regneri2010learning, li2012crowdsourcing, wanzare2016crowdsourced, rashkin-etal-2018-event2mind} and are hence small. 
Existing work has also leveraged wikiHow for large-scale knowledge-base construction \cite{jung2010automatic, chu2017distilling,park2018learning}, but our work is the first to provide a comprehensive intrinsic and extrinsic evaluation of the resulting knowledge-base. }

\section{Conclusion}
We propose a search-then-rerank algorithm to effectively construct a hierarchical knowledge-base of procedures based on wikiHow. 
Our hierarchies are shown to help users accomplish tasks by accurately providing decomposition of a step and improve the performance of downstream tasks such as retrieving instructional videos. 
\newsz{One interesting extension is to further study and improve the robustness of our two-stage method to tackle more complex linguistic structures of steps and goals~(\eg~negation, conjunction). Another direction is to enrich the resulting knowledge-base by applying our method to other web resources,\footnote{\eg~\url{https://www.instructables.com/}, \url{https://www.diynetwork.com/how-to}}} or to other modalities~(\eg~video clips).
Future work could also explore other usages such as comparing and clustering procedures based on their deep hierarchies; or applying the procedural knowledge to control robots in the situated environments.

\section*{Acknowledgments}
This research is based upon work supported in part by the DARPA KAIROS Program (contract FA8750-19-2-1004), the DARPA LwLL Program (contract FA8750-19-2-0201), the IARPA BETTER Program (contract 2019-19051600004), and the Amazon Alexa Prize TaskBot Competition. Approved for Public Release, Distribution Unlimited. The U.S. Government is authorized to reproduce and distribute reprints for Governmental purposes notwithstanding any copyright notation thereon. The views and conclusions contained herein are those of the authors and should not be interpreted as necessarily representing the official policies, either expressed or implied, of Amazon, DARPA, IARPA, or the U.S. Government.

We thank Ziyang Li and Ricardo Gonzalez for developing the web demo, John Wieting for support on implementation, and the anonymous crowd workers for their annotations. 


\bibliography{anthology,custom}
\bibliographystyle{acl_natbib}
\clearpage
\appendix
\section{Crowdsourcing Details}\label{sec:cs_details}

As discussed in \autoref{sec:human_eval}, we use Amazon Mechanical Turk (mTurk) to collect human judgements of linked wikiHow articles. Our mTurk task design HTML is attached in the supplementary materials. Each task includes an overview, examples of ratings, and 11 questions including 1 control question. Each question has the following prompt:

\begin{displayquote}
Imagine you're reading an article about the goal \texttt{c\_goal}, which includes a step \texttt{step}. Then, you're presented with a new article \texttt{r\_goal}. Does this new article help explain how to do the step \texttt{step}?
\end{displayquote}
where \texttt{c\_goal} is the original corresponding goal of the \texttt{step}, and \texttt{r\_goal} is the retrieved goal by the model. Both \texttt{c\_goal} and \texttt{r\_goal} have hyperlinks to the wikiHow article. The options of rating are:
\begin{enumerate}[itemsep=0mm]
 \item The article explains exactly how to do the step.
 \item The article is helpful, but it either doesn't have enough information or has too much unrelated information.
 \item The article explains something related, but I don't think I can do the step with the instructions.
 \item The article is unhelpful/unrelated.
 \item I don't know which option to choose, because: [text entry box]
\end{enumerate}

The control question contains either a \texttt{step} and \texttt{r\_goal} with the exact same texts once lower-cased (in which case the expected answer is always \#1), or a \texttt{step} and a randomly selected unrelated \texttt{r\_goal} (in which case the expected answer is always \#4). We estimate that answering each question would take 30 seconds, with a pay of \$0.83 per task which equates to an hourly rate of \$9.05. We require workers to be English-speaking, with the mTurk Master qualification and a lifetime approval rate of over 90\%. 



To sample examples to annotate, we first obtain all the steps corresponding to the same 1000 goals as we did in \autoref{sec:dataset_construction}. To evaluate the \unlinkable's ability to predict \texttt{unlinkable}, we randomly sample 500 steps predicted as \texttt{unlinkable} and another 500 predicted as otherwise. Then, for these 1000 steps, we obtain linked goal predictions of our three models: \unlinkable, \deberta, and the SP model. If \unlinkable predicts a step to be \texttt{unlinkable} by ranking the placeholder token first, the second ranked goal is instead considered. After removing duplicates of predicted step-goal pairs, we are left with 1448 examples. 

When performing analyses, we only consider the responses from crowdworkers that pass more control questions than they fail.

\begin{figure*}[t]
    \centering
    \includegraphics[width=\textwidth]{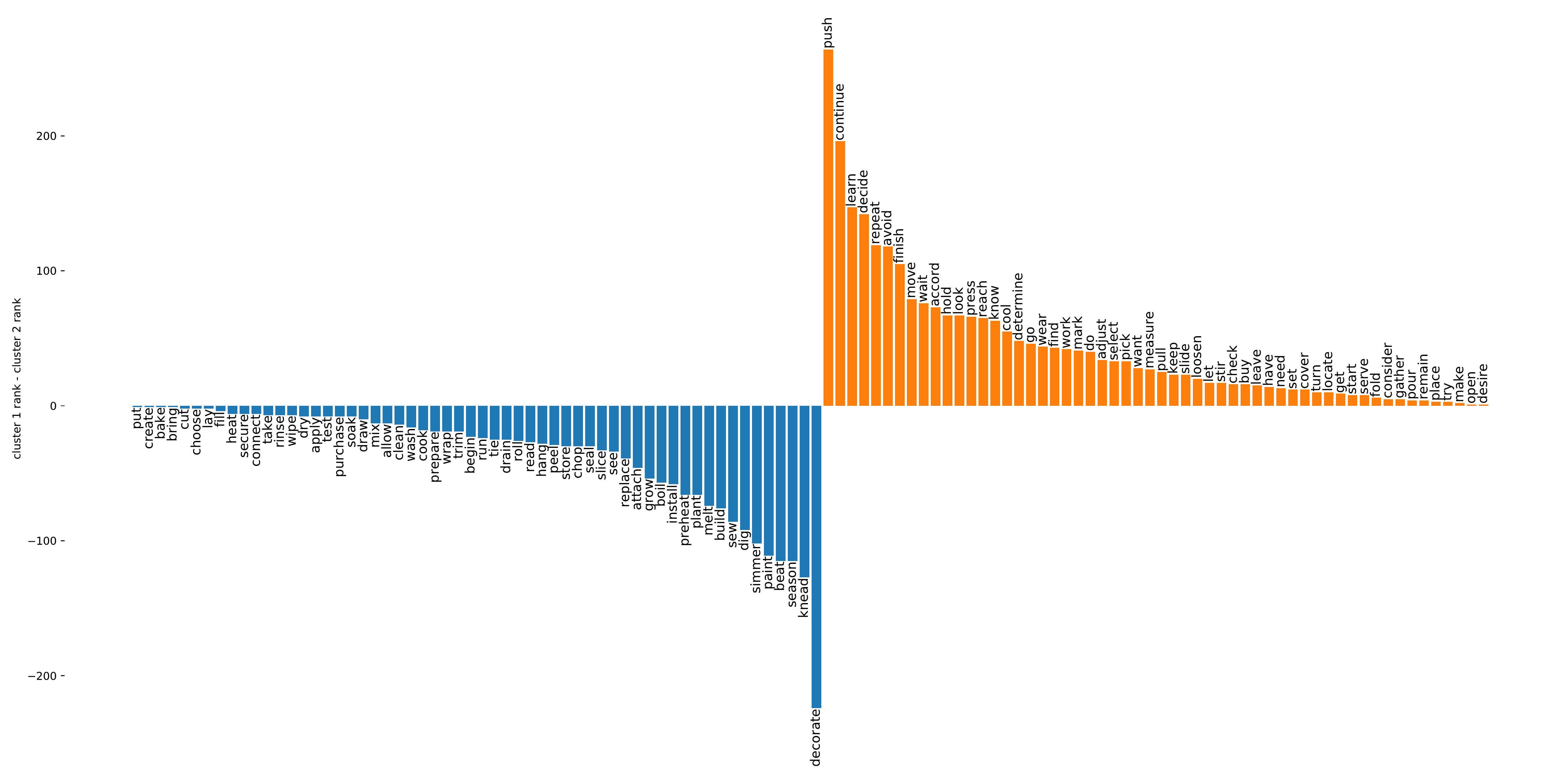}
    \caption{The full version of \autoref{fig:part_verb_distribution}}
    \label{fig:full_verb_distribution}
\end{figure*}

\RestyleAlgo{ruled}
\begin{algorithm}[t]
\small
\caption{Video-based filtering}\label{alg:search}
\KwData{goal $g$, cost function $f$,
candidate steps $\bm{p}=[p_1, ..., p_n]$, relevant videos $\vv_{g}^\textrm{tr}$\\}
\KwResult{$best\_query$}
$k \gets 15$\;
$best\_query \gets [g]$\;
$min\_cost \gets f(best\_query, \vv_{g}^\textrm{tr})$\;
$r \gets \min(n, k)$\;
\While{$r \geq 0$}{
    $in\_cost \gets 1e10$\;
    \For{$p$ in $\bm{p}$}
    {
        \If{$p$ not in $best\_state$}
        {
        $query \gets [best\_query, p]$\;
        $cost \gets f(query, \vv_{g}^\textrm{tr})$\;
        \If{$cost < in\_cost$}
            {
                $in\_cost \gets cost$\;
                $in\_query \gets query$\;
            }
        }
    }
    \uIf{$in\_cost < min\_cost$}{
      $min\_cost \gets in\_cost$\;
      $best\_query \gets in\_query$\;}
    \uElse{\textit{break}}
  $r = r - 1$\;
}
\end{algorithm}

\section{Video Retrieval Setup}
\subsection{Dataset Construction}\label{sec:vr_split}
Existing works also practice similar data splits that share the labels of videos/images across the training, development and the test set. For example, image retrieval tasks use the same objects labels for training and evaluations~\cite{wan2014deep}; Activity Net~\cite{caba2015activitynet}, a popular benchmark for human activity understanding, uses the same 203 activities across different splits; \citet{yang-etal-2021-visual} trains a step inference model with a training set that shares the same goals with the test set. 

This data split is meaningful on its own. We can view the original queries as initial schemas for complex procedures. Then we induce more \emph{generalizable} schemas by matching them with schema instantiations (in our case, the videos that display the procedures). We evaluate the quality of the induced schemas by matching them with \emph{unseen} instantiations. The large-scale DARPA KAIROS project\footnote{\url{https://www.darpa.mil/program/knowledge-directed-artificial-intelligence-reasoning-over-schemas}} adopted a similar setup, which we believe indicates its great interest to the community. 

In terms of the scale of the video retrieval dataset, though we only select 1000 goals from 23$k$ goals from Howto1M, there are already ~150$k$ videos in total while widely-used video datasets like COIN~\cite{tang2019coin} only contain 180 goals and ~10$k$ videos. In addition, exiting works like \cite{yang-etal-2021-visual} also experimented with a sampled dataset of similar scale.

\subsection{Evaluation Metrics}\label{sec:eval_metrics}
We report precision@$N$, recall@$N$ and mean rank (MR) following existing works on video retrieval~\cite{Luo2021CLIP4Clip}
\begin{align}
\small
\label{eqn:eqlabel}
\begin{split}
 \texttt{recall@N}&=\frac{1}{M}\sum_{i=1}^{M} \frac{\sum_{v_{j} \in {\vv_{g_i}}} \mathbb{1}(r(v_j)<=N)}{|\vv_{g_i}|}
\\
  \texttt{precision@N}&=\frac{1}{M}\sum_{i=1}^{M} \frac{\sum_{v_{j} \in {\vv_{g_i}}} \mathbb{1}(r(v_j)<=N)}{N}
\\
  \texttt{MR}&=\frac{1}{M}\sum_{i=1}^{M} \frac{\sum_{v_{j} \in {\vv_{g_i}}} r(v_j)}{|\vv_{g_i}|}
\end{split}
\end{align}
where $M$ is the number of goals in total, $\vv_{g_i}$ is a set of ground truth videos of goal $g_i$ is the rank of video $v$ and $\mathbb{1}$ is the indicator function.

\section{Experiment Reproducibility}
\paragraph{Candidate Goal Retrieval}The detailed parameter information of \textsc{SP} can be found in S5.1 in~\cite{wieting2021paraphrastic}. Encoding all steps and goals in wikiHow took around two hours on a 2080Ti (12GB) GPU. For \textsc{Sbert}, the encoding took around an hour on a v100 GPU (32GB). 
\paragraph{Reranking} We used the \texttt{transformers} library~\cite{wolf2020transformers} for re-ranking. The two re-ranking models we used are ``bert-base-uncased'' and ``deberta-v2-large-mnli''. We finetuned each model on our training set for five epochs and selected the best model on the validation set. Finetuning took around two hours on a 2080Ti (12GB) GPU for BERT and eight hours on a v100 GPU (32GB) for \textsc{DeBERTa}. We used the default hyper-parameters provided by the \texttt{transformers} library. 

\section{Risks}
Our resulting hierarchy contains events from wikiHow, which may contain unsafe content that slip through its editorial process, although this is relatively unlikely.

\section{License of Used Assets}
The wikiHow texts used in this work are licensed under CC BY-NC-SA 3.0. \\
FAISS is licensed under MIT License. \\
BERT is licensed under Apache License 2.0.\\
DeBERTa is licensed under MIT License. \\
The SP model is licensed under BSD 3-Clause "New" or "Revised" License
ElasticSearch is licensed under Apache License 2.0.\\
HowTo100M is licensed under Apache License 2.0.\\
\end{document}